\documentclass[journal]{IEEEtran}

\usepackage{amsmath,amssymb,amsfonts}
\usepackage{booktabs,tabularx,multirow}
\usepackage{algorithmic}
\usepackage{graphicx}
\usepackage[numbers]{natbib}
\usepackage{colortbl}

\usepackage{makecell}
\usepackage{textcomp}
\usepackage{xcolor}
\usepackage{caption} 
\usepackage{listings}
\usepackage{subcaption}

\usepackage{ragged2e}
\newcolumntype{Y}{>{\RaggedRight\arraybackslash}X}
\newcommand{\NA}{\multicolumn{1}{c}{\textemdash}}

\def\BibTeX{{\rm B\kern-.05em{\sc i\kern-.025em b}\kern-.08em
    T\kern-.1667em\lower.7ex\hbox{E}\kern-.125emX}}

\definecolor{codegreen}{rgb}{0,0.6,0}
\definecolor{codegray}{rgb}{0.5,0.5,0.5}
\definecolor{codepurple}{rgb}{0.58,0,0.82}
\definecolor{backcolour}{rgb}{0.98,0.98,0.98}

\newcolumntype{Y}{                                                                                                                                                  >{\raggedright\arraybackslash}X}
\renewcommand{\arraystretch}{1.15}

\title{Accuracy, Memory Efficiency and Generalization: A Comparative Study on Liquid Neural Networks and Recurrent Neural Networks}

\author{\IEEEauthorblockN{Shilong Zong}
\and
\IEEEauthorblockN{Alex Bierly}
\and
\IEEEauthorblockN{Almuatazbellah Boker}
\and
\IEEEauthorblockN{Hoda Eldardiry}
}

\date{}

\begin{document}
\maketitle

\begin{abstract}
    This review aims to conduct a comparative analysis of liquid neural networks (LNNs) and traditional recurrent neural networks (RNNs) and their variants, such as long short-term memory networks (LSTMs) and gated recurrent units (GRUs). The core dimensions of the analysis include model accuracy, memory efficiency, and generalization ability. By systematically reviewing existing research, this paper explores the basic principles, mathematical models, key characteristics, and inherent challenges of these neural network architectures in processing sequential data. Research findings reveal that LNN, as an emerging, biologically inspired, continuous-time dynamic neural network, demonstrates significant potential in handling noisy, non-stationary data, and achieving out-of-distribution (OOD) generalization. Additionally, some LNN variants outperform traditional RNN in terms of parameter efficiency and computational speed. However, RNN remains a cornerstone in sequence modeling due to its mature ecosystem and successful applications across various tasks. This review identifies the commonalities and differences between LNNs and RNNs, summarizes their respective shortcomings and challenges, and points out valuable directions for future research, particularly emphasizing the importance of improving the scalability of LNNs to promote their application in broader and more complex scenarios.
\end{abstract}

\section{Introduction}
    \IEEEPARstart{S}{equence} modeling plays a crucial role in numerous fields of artificial intelligence, such as natural language processing, speech recognition, time series prediction, and robot control \cite{lawrynczuk2021lstm}, \cite{chen2024graph}. As the complexity, dynamism, and noise interference of real-world data continue to increase, there is a growing demand in both academia and industry for sequence models that are not only accurate but also efficient and robust.

\renewcommand{\thefootnote}{}  
\footnotetext{%
Shilong Zong is with the Department of Computer Science, Virginia Tech, Blacksburg, VA 24061 USA (e-mail: shilongz@vt.edu). \par
Alex Bierly is with the Department of Electrical and Computer Engineering, Virginia Tech, Blacksburg, VA 24061 USA (e-mail: akbierly@vt.edu). \par
Almuatazbellah Boker is with the Department of Electrical and Computer Engineering, Virginia Tech, Blacksburg, VA 24061 USA (e-mail: boker@vt.edu). \par
Hoda Eldardiry is with the Department of Computer Science, Virginia Tech, Blacksburg, VA 24061 USA (e-mail: hdardiry@vt.edu).%
}    

    Recurrent neural networks (RNNs) and their important variants, such as long short-term memory (LSTM)  \cite{lawrynczuk2021lstm} and gated recurrent units (GRUs), are the foundational architectures for deep learning in sequence data processing. These models capture temporal dependencies through their internal recurrent connections and memory units, achieving significant success across various tasks. However, traditional RNNs also face well-known limitations, including difficulty in effectively learning very long-range dependencies \cite{li2021curse}, potential issues such as gradient vanishing or exploding during training, and inherent computational efficiency bottlenecks when handling extremely long sequences.
    
    In recent years, Liquid Neural Networks (LNNs) have emerged as a novel category of neural networks, attracting significant attention. LNNs draw inspiration from biological neural systems (such as the nervous system of the nematode Caenorhabditis elegans) \cite{hasani2021ltc}, \cite{chahine2023robust}, \cite{kaddoura2024exploring}, \cite{lechner2020neural} and continuous-time dynamic systems theory. Unlike traditional RNNs, which operate on discrete time steps, LNNs describe the continuous evolution of their neural states through ordinary differential equations (ODEs) \cite{hasani2021ltc}, \cite{treven2023efficient}. This fundamental difference enables LNNs to adaptively adjust their behavior and temporal scales according to the dynamic characteristics of input data, thereby potentially overcoming some inherent limitations of RNNs, particularly in handling irregularly sampled data, noise interference, and achieving stronger generalization capabilities.
    
    Many real-world phenomena are inherently continuous, and LNN's continuous-time dynamic properties enable it to naturally represent time-varying signals and potentially handle irregularly sampled data more effectively. The core innovation of LNN lies in its time-processing mechanism, which may offer inherent advantages in specific scenarios. The limitations of RNN in handling long-range dependencies and gradient issues have directly driven the exploration of alternative architectures like LNNs \cite{li2021curse}. 
 
    This review aims to conduct a comprehensive comparative study of LNN and traditional RNN in terms of model architecture, mathematical foundations, accuracy, memory efficiency, and generalization ability. This paper will systematically review the relevant literature, identify their commonalities and differences, summarize their respective shortcomings and challenges, and point out valuable directions for future research, with a particular focus on the scalability of LNN. While existing literature includes comparisons of specific models on particular tasks \cite{hasani2021ltc}, \cite{chahine2023robust}, \cite{hasani2022cfc}, a comprehensive review that systematically contrasts the LNN and RNN architectural families across the core dimensions of accuracy, memory efficiency, and generalization remains less common \cite{kaddoura2024exploring}, \cite{karn2024generalized}. This paper aims to fill this gap by synthesizing recent advancements to provide a holistic perspective, with a particular emphasis on the emerging LNN paradigm and its potential to address the inherent limitations of traditional recurrent models.

    Overall, the contribution of this paper is manifested in providing a comparative study of recurrent neural networks (RNNs) and Liquid Neural Networks (LNNs) and illustrating the advantages of LNNs for predicting data with long-term time dependencies. We support our study with three case studies; namely, trajectory prediction task using real-world motion capture data, a synthetic time series prediction task involving damped sine waves, and modeling Intensive Care Unit (ICU) patient state evolution. Finally, we outline future research directions for LNNs.
    
    The structure of this paper is as follows: Section 3 provides a detailed introduction to the model architecture and theoretical basis of recurrent neural networks (RNNs) and Liquid Neural Networks (LNNs). Section 4 compares and analyzes the two models in terms of accuracy, memory efficiency, and generalization ability. Section 5 presents a more practical case study. Section 6 discusses future research directions and open issues. Section 7 summarizes the entire paper.

\section{Model Architecture}
\begin{figure}[h]
        \centering
        \includegraphics[width=\linewidth]{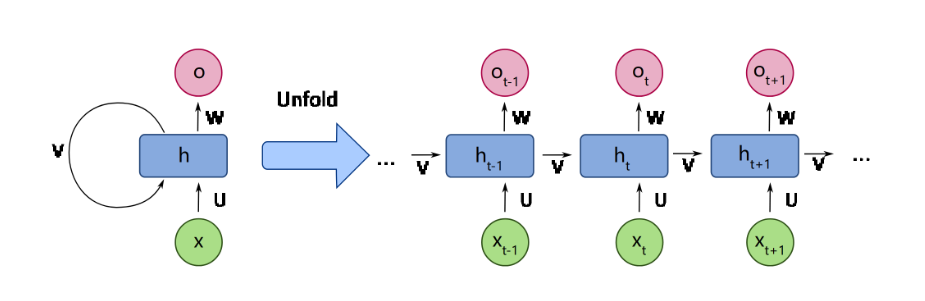}
        \caption{Recurrent Neural Network (RNN) unfolding representation illustrating the temporal expansion of recurrent connections. The left side shows the compact recurrent structure with feedback connections, while the right side demonstrates the unfolded network across multiple time steps. Each time step receives input \(x_t\), updates hidden state \(h_t\), and produces output \(o_t\), with weight matrices (W, U, V) shared across all time steps.}
        \label{fig:RNN}
    \end{figure}

\begin{figure}[h]
        \centering
        \includegraphics[width=\linewidth]{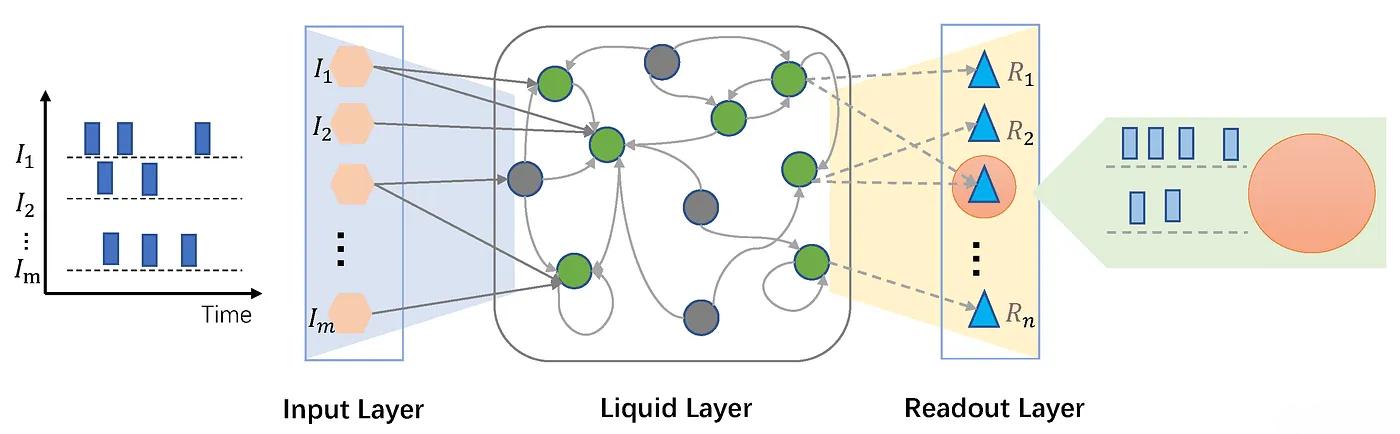}
        \caption{Conceptual architecture of a Liquid Neural Network (LNN). 
            \textbf{Input layer (left):} $m$ input streams $I_1,\ldots,I_m$ provide event-like signals over time, forming the input vector $x(t)$.
            \textbf{Liquid layer (middle):} a recurrent, heterogeneous reservoir with continuous-time state $h(t)$; its dynamics follow an ODE $\dot h(t)=f\!\big(h(t),x(t);\theta\big)$, and the arrows depict recurrent couplings among units. 
            \textbf{Readout layer (right):} $n$ readout units $R_1,\ldots,R_n$ compute task-specific outputs from the liquid state, typically 
            $r(t)=W h(t)+b$; dashed lines indicate dense projections from the liquid layer to each readout. 
            The far-right sketch illustrates example output trajectories/decisions produced by the readouts.}
        \label{fig:LNN}
    \end{figure}

    To understand the core differences and potential of LNNs and RNNs, we first need to analyze their respective architectural designs and mathematical principles in depth. RNNs and their gated variants capture sequence dependencies through iterative updates at discrete time steps, while LNNs introduce continuous-time dynamics, which fundamentally change their behavior. To illustrate the fundamental architectural differences, Figures~\ref{fig:RNN} and~\ref{fig:LNN} provide a conceptual comparison between the discrete-time processing of RNNs and the continuous-time dynamics of LNNs. The subsequent Table~\ref{tab:overview-rnn-lnn} summarizes the main characteristics of the RNN and LNN model families.
    
\begin{table*}[!t]
  \caption{Overview of Recurrent Neural Network (RNN) and Liquid Neural Network (LNN) Model Families}
  \label{tab:overview-rnn-lnn}
  \centering
  \footnotesize
  \begin{tabularx}{\textwidth}{@{}l l Y Y Y Y@{}}
    \toprule
    \textbf{Family} & \textbf{Architecture} & \textbf{Core Principle} & \textbf{Temporal Mechanism} & \textbf{Advantages} & \textbf{Limitations} \\
    \midrule
    RNN & Standard RNN & Discrete recurrence & Sequential hidden-state update & Simple structure; processes sequences & Vanishing/exploding gradients; struggles with long-term deps \\
    RNN & LSTM & Gated discrete recurrence & Cell state with forget / input / output gates & Mitigates gradient issues; captures long-term deps & Many parameters; computationally heavy \\
    RNN & GRU & Simplified gated recurrence & Update and reset gates & Fewer params; faster than LSTM; similar accuracy & Slightly less expressive on some tasks \\
    \midrule
    LNN & LTC & Continuous ODE with dynamic $\tau$ & NN-modulated linear ODE & Bounded stability; adaptive scales & Needs ODE solver; stiff eqs hard \\
    LNN & CfC & Closed-form ODE approximation & Embedded NN combination & Solver-free; fast training/inference & Approximation error; theory complex \\
    LNN & NCP & Sparse bio-inspired structure & ODE neurons, sparse connectivity & Compact; interpretable; robust & Sparse design needs expert tuning \\
    LNN & Liquid-S4 & Linearised LTC state-space model & Linearised LTC state evolution & Excellent long-range deps; parameter-efficient & Relies on SSM theory; design complex \\
    LNN & LRC / LRCU & ODE with liquid capacitance & State-dependent capacitance; Euler update & Improves LTC oscillations; efficient & New approach; ecosystem immature \\
    \bottomrule
  \end{tabularx}
\end{table*}

    \subsection{Recurrent Neural Network}
        RNN is a neural network designed for processing sequential data. Its core idea is to use internal recursive structures to transmit and maintain information from previous time steps.
        \subsubsection{Standard RNN}
            Standard RNNs process each element of an input sequence in order. At each time step, the network receives the current input and the hidden state from the previous time step, then calculates a new hidden state and the current output. The hidden state acts as a compressed representation of past information for the network.\\
            The core calculation process of a standard RNN can be described by the following formula:\\
            Hidden status updates:
            \begin{align}
                h_t = \sigma_h\bigl(W_{hh}h_{t-1} + W_{xh}x_t + b_h\bigr),
            \end{align}
            Output calculation:
            \begin{align}
                y_t = \sigma_y\bigl(W_{hy}h_t + b_y\bigr),
            \end{align}
            where \(x_t\) is the input at time step \(t\), \(h_t\) is the hidden state, \(h_{t-1}\) is the hidden state at the previous time step, and \(y_t\) is the output. \(W_{xh}\), \(W_{hh}\), and \(W_{hy}\) are the weight matrices from the input to the hidden layer, hidden layer to hidden layer, and hidden layer to output layer, respectively. \(b_h\) and \(b_y\) are bias vectors. \(\sigma_h\) and \(\sigma_y\) are typically activation functions, such as tanh or sigmoid.

        \subsubsection{Long Short-Term Memory (LSTM) Network}
            LSTM was proposed to solve the gradient vanishing or explosion problem faced by standard RNNs when learning long-range dependencies \cite{lawrynczuk2021lstm}. LSTM introduces an explicit memory unit (cell state \(C_t\)) and three gating mechanisms—forget gate, input gate, and output gate—to finely control the flow of information in the cell state.\\
            The gating mechanism and state update of LSTM are defined by the following formulas:\\
            Forgetfulness Gate (\(f_t\)):
            \begin{align}
                f_t = \sigma\bigl(W_f\,[h_{t-1},\,x_t] + b_f\bigr).
            \end{align}
            This gate decides which information to discard from the cellular state.
            
            Input Gate ($i_t$) and Candidate Cell State ($\widetilde C_t$):
            
            This stage determines which new information will be stored in the cell state. It comprises two parts working in tandem: the input gate ($i_t$), which uses a sigmoid function to decide which values will be updated, and a tanh layer that generates a vector of new candidate values, the candidate cell state ($\widetilde C_t$).
            \begin{align}
                i_t &= \sigma\bigl(W_i\,[h_{t-1},\,x_t] + b_i\bigr),\\
                \widetilde C_t &= \tanh\bigl(W_c\,[h_{t-1},\,x_t] + b_c\bigr).
            \end{align}
            
            This gate decides which new information to store in the cell state. It consists of two parts: a sigmoid layer which decides the update proportion (the input gate \(i_t\)), and a tanh layer that creates a vector of new candidate values, \(\widetilde C_t\), which could be added to the cell state.
          
            Cell State Update ($C_t$): 
            \begin{align}
                C_t = f_t \,\odot\, C_{t-1} \;+\; i_t \,\odot\, \widetilde C_t,
            \end{align}
            This gate combines the forget gate and the input gate to update the cell state.
            Output Gate ($o_t$): 
            \begin{align}
                o_t = \sigma\bigl(W_o\,[h_{t-1},\,x_t] + b_o\bigr),
            \end{align}
            This gate decides which parts of the cell state to output.
            Hidden State Update ($h_t$):
            \begin{align}
                h_t = o_t \,\odot\, \tanh\bigl(C_t\bigr),
            \end{align}
            Here,$W_f, W_i, W_c, W_o$ are weight matrices, $b_f, b_i, b_c, b_o$ are bias vectors, $\sigma(\cdot)$ is the sigmoid function, $\tanh(\cdot)$ is the hyperbolic tangent function, $\odot$ denotes element‐wise multiplication, and $[h_{t-1},\,x_t]$ denotes the concatenation of $h_{t-1}$ and $x_t$.

        \subsubsection{Gate-controlled Recirculation Unit}
            GRU is a simplified version of LSTM, designed to maintain performance comparable to LSTM while reducing the number of parameters and computational complexity. GRU merges the forget gate and input gate of LSTM into a single “update gate” and directly fuses the cell state and hidden state.\\
            The core equation of GRU is as follows:\\
            Reset Gate ($r_t$): Determines how much of the past information to forget.
            \begin{align}
                r_t = \sigma\bigl(W_r\,[h_{t-1},\,x_t] + b_r\bigr),
            \end{align}
            Update Gate ($z_t$): Decides how much of the past hidden state and how much of the candidate hidden state to use.
            \begin{align}
                z_t = \sigma\bigl(W_z\,[h_{t-1},\,x_t] + b_z\bigr),
            \end{align}
            Candidate Hidden State ($\tilde h_t$): Computes the candidate activation for the current time step.
            \begin{align}
                \tilde h_t = \tanh\bigl(W_h\,[\,r_t \odot h_{t-1},\;x_t] + b_h\bigr)
            \end{align}
            Final Hidden State ($h_t$): Combines the previous hidden state and the candidate hidden state according to the update gate.
            \begin{align}
                h_t = (1 - z_t)\,\odot\,h_{t-1} \;+\; z_t\,\odot\,\tilde h_t,
            \end{align}
            where, $W_r, W_z, W_h$ are weight matrices, $b_r, b_z, b_h$ are bias vectors.
            
        \subsubsection{Intrinsic challenges of RNNs}
            Although LSTM and GRU alleviate the problems of standard RNNs to some extent through gating mechanisms, they still face some inherent challenges:

            \begin{enumerate}
                \item Gradient vanishing/explosion: When processing very long sequences, even LSTM and GRU may experience gradients that become too small or too large during backpropagation through time (BPTT), thereby hindering effective learning~\cite{li2021curse}.
                \item Capturing extremely long-range dependencies: Although LSTM/GRU are designed to capture long-range dependencies, their ability to capture dependencies in extremely long sequences (e.g., thousands of time steps) remains limited in practice \cite{hasani2021ltc}.
                
                \item Computational cost: The sequential processing nature of RNNs makes them difficult to parallelize at scale like convolutional networks or transformers, which may result in slower training and inference speeds when handling very long sequences. Additionally, training RNNs may require significant computational resources and memory.
            \end{enumerate}
            
            The evolution from standard RNNs to LSTM/GRU was primarily driven by the need to address gradient issues and improve memory capabilities. This backdrop laid the groundwork for the emergence of LNNs, which aim to tackle these challenges from a fundamentally different angle—continuous dynamics rather than discrete gating.

    \subsection{Continuous-Time Neural Networks and the Liquid Neural Networks (LNNs) Family}
    
        Continuous-time neural networks represent a class of models where the evolution of neuron states is described by ordinary differential equations (ODEs), rather than discrete-time recurrence relations \cite{chen2018node}. Liquid Neural Networks (LNNs) are a prominent and biologically-inspired subclass of these models \cite{hasani2021ltc}, \cite{treven2023efficient}. A key characteristic of many LNN variants is their use of state- and input-dependent dynamics, often realized through learned, adaptive time constants, which allows them to dynamically adjust their response properties to incoming signals. Unlike RNNs, which operate on discrete time steps, the state of a continuous-time model is a continuous function of time. This continuity enables them to naturally handle irregularly sampled time series data and model underlying continuous processes. Specifically, LNNs are a biologically-inspired subset of these models, often inspired by the neural circuitry of the nematode Caenorhabditis elegans \cite{hasani2021ltc}, \cite{lechner2020neural}. A key feature of many LNN variants is that their neurons can dynamically adjust their response time or “memory span” based on input signals, often through a learned time constant. The LNN family has evolved rapidly, leading to several key architectures with distinct trade-offs in performance, computational efficiency, and interpretability. The following subsections will detail these foundational variants, from the original Liquid Time-Constant (LTC) networks to more recent, specialized designs.
        
 
        \subsubsection{Basic LNN}
            
            The dynamics of LNN are usually described by a set of ordinary differential equations, which can be written in general form as:
            \begin{align}
                \frac{d h(t)}{d t} = f\bigl(h(t),\,x(t),\,t,\,\theta\bigr),\label{1}
            \end{align}
            In this context, \(h(t)\) represents the hidden state vector of the network at time \(t\), \(x(t)\) is the input vector, and \(\theta\) is the learnable parameter of the network. The function $f$ is a general nonlinear function, parameterized by $\theta$, that defines the dynamics of the hidden state. In practice, $f$ is typically implemented as a neural network, such as a multi-layer perceptron (MLP), which takes the current state $h(t)$ and input $x(t)$ as its inputs to compute the state's rate of change.
            
            When \eqref{1} has no analytical solution (i.e. no closed-form solution), numerical ODE solvers (such as Runge-Kutta methods, DOPRI5, or Euler methods) are used to approximate the system's state evolution at discrete time points.\\
            
            The LNN family has evolved rapidly, showing a clear trend from general-purpose models requiring numerical solvers to more efficient and specialized architectures. This progression begins with foundational models like neural ODEs and the original Liquid Time-Constant (LTC) networks \cite{hasani2021ltc}, \cite{treven2023efficient}. To address the computational cost of solvers, Closed-form Continuous-time (CfC) networks \cite{hasani2022cfc} were developed to provide an analytical approximation. Concurrently, Neural Circuit Policies (NCPs) \cite{lechner2020neural} emphasized sparsity and biological interpretability. More recent advancements include hybrid models like Liquid-S4 \cite{hasani2023liquid}, which integrates LNN principles with powerful State-Space Models (SSMs), and Liquid Resistance-Capacitance (LRC) networks \cite{farsang2024liquid}, which refine the core ODE mechanics for improved stability and biological plausibility. This evolution highlights a consistent research direction toward making LNNs more practical, powerful, and versatile. The following subsections will detail these key variants, each defined by a distinct mathematical formulation.
            
        \subsubsection{Liquid time constant network (LTC)}
            LTC is a specific type of LNN whose core idea is that the “time constant” of neurons is dynamic and learned by the network itself based on input and current state\cite{hasani2021ltc}. Its state equation is typically expressed as:
            \begin{equation}
              \begin{split}
                \frac{dx(t)}{dt}
                &= -\Bigl(\tfrac{1}{\tau} + \mathrm{NN}(x(t),I(t),\theta)\Bigr)\odot x(t) \\
                &+ \mathrm{NN}(x(t),I(t),\theta)\odot A
              \end{split}\label{ltc}
            \end{equation}
            where \(x(t)\) is the hidden state, \(I(t)\) is the input, \(\tau\) is a base time-constant vector, and A is a learnable bias vector. The term $\mathrm{NN}(\cdot)$ represents a parameterized nonlinear mapping that modulates the system's dynamics based on the current state and input. This mapping is typically implemented as a shallow neural network with a sigmoid or tanh activation function. Its output dynamically adjusts both the decay rate of the state (i.e., the effective time constant) and its coupling to the bias term A.
            
            The core difference between the update mechanisms of LTC and RNN lies not in trainable vs. fixed weights, but in how those weights define the system's dynamics. In an RNN, the weight matrices like \(W_{hh}\) and \(W_{xh}\) are fixed parameters that define a static state-transition function. In contrast, the parameters within the LTC's \(\mathrm{NN}(\cdot)\) function are also fixed after training. However, the output of this function, which dynamically modulates the ODE's coefficients (e.g., the effective time constant), changes at every moment based on the current state \(x(t)\) and input \(I(t)\). This makes the system's temporal dynamics inherently adaptive and state-dependent, allowing neurons to adjust their response and memory characteristics on-the-fly. This adaptive property is a key mechanism for handling non-stationary data and is central to LTC's reported robustness and generalization capabilities.
            
            In RNNs, \(W_{hh}\) and \(W_{xh}\) are fixed during the inference stage after learning. In LTCs, however, the terms \(\frac{1}{\tau} + \mathrm{NN}\bigl(x(t),I(t),\theta\bigr)\) act as the reciprocal of an effective time constant, which varies with the current state \(x(t)\) and input \(I(t)\) because NN is a neural network. This means that the rate at which neurons “forget” or “respond” is not static but adaptive. This adaptive property embedded in the basic ODE is a powerful mechanism for handling changing temporal patterns and data non-stationarity, which may be the key reason for its reported robustness and generalization ability.
            
            LTC exhibits stable and bounded behavior and has good expressive capabilities. To address the stiff ordinary differential equations (ODEs) often encountered in LTCs, a practical fixed-step ODE solver known as the "Fused Solver" was introduced. This solver is designed to combine the stability of implicit Euler methods with the efficiency of their explicit counterparts \cite{hasani2021ltc}.
            
        \subsubsection{Closed-form solutions for continuous-time neural networks (CfC)}
            The primary motivation for CfC \cite{hasani2022cfc} is to avoid the high computational cost and complexity associated with numerical ODE solvers in models like LTC. This is based on the assumption that the differential equation appears in a linear form. Its final form is typically represented as: 
            \begin{equation}
                \begin{split}
                    x(t) &= \sigma\bigl(-\,f(x,\,I;\,\theta_{f})\,t\bigr)\;\odot\;g(x,\,I;\,\theta_{g})\\
                    &\quad + \bigl[1 - \sigma\bigl(-\,f(x,\,I;\,\theta_{f})\,t\bigr)\bigr]\;\odot\;h(x,\,I;\,\theta_{g})\,,
                \end{split}
            \end{equation}
            where $f$, $g$, and $h$ are nonlinear functions parameterized by learnable weights ($\theta_{f}$, $\theta_{g}$, and $\theta_{h}$), which are typically implemented as shallow neural networks. The function \(\sigma\) is the sigmoid function, acting as a mixing gate. Compared with solver-based LNNs, CfC has faster training and inference speeds and a smaller computational footprint.

        \subsubsection{Neural Circuit Policy (NCP)}
            NCP refers to LNNs with sparse, biologically inspired connection structures, typically constructed using LTC or CfC neurons \cite{lechner2020neural}. NCPs typically adopt a four-layer design (sensory layer, intermediate layer, command layer, and motor layer). NCPs emphasize model compactness, interpretability, and robustness.

        \subsubsection{Liquid-S4 (LTC State Space Model)}
            The standard continuous-time state-space model (SSM) is expressed as $x'(t)=Ax(t)+Bu(t),~y(t)=Cx(t)+Du(t)$. Recently proposed sequence-structured state-space models (S4) and the Mamba model have introduced structural and selective improvements, demonstrating outstanding performance in long-sequence modeling \cite {gu2021s4}, \cite{gu2023mamba}, \cite{gu2022hippo}.

            Liquid-S4 integrates the principles of Long-Term Modeling (LTC) with the Structured State Space Model (SSM), aiming to balance LTC's generalization capability with S4's scalability for handling long sequences. Its dynamic equations can be expressed as \cite{hasani2023liquid}:
            \begin{align}
                \dot{x} &= (A + B\,u)\,x + B\,u,\\
                y       &= C\,x,
            \end{align}
            where $A,B$ and $C$ are matrices of appropriate dimensions. 
            The LTC state space model demonstrates improved generalization capabilities on long-range dependency tasks and typically requires fewer parameters than the S4 model.
            
        \subsubsection{Liquid Resistive Neural Network (LRC)}
            The LRC network is an extension of the LTC and saturated liquid time constant STC networks (Saturated Liquid Time-Constant (STC)), introducing a “liquid capacitance” term to make the membrane capacitance state-dependent, aiming to enhance biological plausibility and suppress oscillations. The LRC unit (LRCU) is its efficient version, solved using an explicit Euler method with single-step expansion.
            
            Compared to LTC/STC, LRC exhibits better generalization, accuracy, and stability, particularly when using simple solvers. Its performance is comparable to that of LSTM, GRU, and neural ODEs \cite{farsang2024liquid}.

\section{Literature-based Comparative Analysis}

\subsection{Accuracy}

    In time series prediction and classification benchmark tests, LTC demonstrates accuracy that is superior to or on par with LSTM, CT-RNN, and neural ODE across various tasks such as gesture recognition, occupancy detection, traffic prediction, and human activity recognition \cite{hasani2021ltc}. For example, in gesture recognition, LTC achieves an accuracy rate of 69.55\%, while LSTM achieves 64.57\% \cite{hasani2021ltc}. In traffic prediction, LTC's mean squared error was 0.099, while LSTM's was 0.169 \cite{hasani2021ltc}. GLNN demonstrated significant accuracy improvements over traditional LNN and neural ODE in tasks such as predicting damped sinusoidal trajectories (GLNN loss 1.0738 vs LNN 2.5494 vs neural ODE 1.9899) and modeling nonlinear RLC circuits (GLNN accuracy 0.95 vs LNN 0.75) \cite{karn2024generalized}. In OCT image analysis, GLNN achieved an accuracy of 0.98 and an F1 score, outperforming traditional LNN (accuracy 0.96, F1 score 0.88) \cite{karn2024generalized}. Liquid-S4 achieved an average performance of 87.32\% on the Long-Range Arena benchmark across image, text, audio, and medical time series, demonstrating state-of-the-art generalization capabilities \cite{hasani2023liquid}. 
    
    On the original speech command dataset, Liquid-S4 achieved an accuracy rate of 96.78\% \cite{hasani2023liquid}. UA-LNN outperformed standard LNN, LSTM, and MLP models in time series prediction, with superior \(R^2\), RMSE, and MAE, and demonstrated higher accuracy, precision, recall, and F1 scores in multi-class classification tasks, especially under noisy conditions \cite{akpinar2025novel}. LRC/LRCU outperforms LSTM, GRU, and MGU in RNN benchmarks and successfully solves neural ODE tasks \cite{farsang2024liquid}. In benchmarks such as PhysioNet, CfC achieves performance comparable to or even better than LSTM and ODE-RNN while being significantly faster \cite{hasani2022cfc}. For traditional RNNS, LSTM and GRU typically exhibit comparable accuracy \cite{lawrynczuk2021lstm}. For complex systems with fewer parameters, LSTM sometimes performs slightly better, but this difference diminishes as the number of neurons increases. 
    
    As shown in Table~\ref{tab:accuracy-benchmarks}, the continuous-time nature and adaptive time constant of LNNs appear to contribute to their strong performance, especially on dynamic or irregularly sampled data that discrete models may struggle with \cite{hasani2021ltc}, \cite{chahine2023robust}. Specific design choices in LNN variants (e.g., the closed-form solution in CfC \cite{hasani2022cfc}, capacitors in LRC \cite{farsang2024liquid}, and uncertainty in UA-LNN \cite{akpinar2025novel}) target specific aspects to improve accuracy in different scenarios. Although many LNN variants claim exceptional accuracy, the specific tasks and conditions where they excel vary. For example, Liquid-S4 performs well on remote dependencies \cite{hasani2023liquid}, UA-LNN performs well under noisy conditions \cite{akpinar2025novel}, and CfC offers a good speed-accuracy trade-off \cite{hasani2022cfc}. This suggests that there is no single "best" LNN, but rather a suite of specialized tools. Different LNNs (LTC, CfC, Liquid-S4, UA-LNN, LRC) have reported high accuracy on different benchmarks or under different conditions. Their architectural innovations are targeted (e.g., CfC for speed, UA-LNN for noise). This means that "high accuracy" in LNNs is not a single property but depends on the specific LNN architecture and the context of the task at hand. Therefore, selecting an LNN requires careful consideration of the problem's characteristics (e.g., sequence length, noise level, computational budget).
    
\begin{table*}[!t]
  \caption{Summary of Accuracy-related Benchmarks Comparing Liquid Neural Networks (LNNs) and Recurrent Neural Networks (RNNs)}
  \label{tab:accuracy-benchmarks}
  \centering
  \footnotesize
  \begin{tabularx}{\textwidth}{@{}l l l Y@{}}
    \toprule
    \textbf{Benchmark / Task} & \textbf{Models} & \textbf{Reported Metric} & \textbf{Key Finding / Comparison} \\
    \midrule
    Gesture recognition & LTC, LSTM & Accuracy (\%) & LTC achieves 69.55\% versus 64.57\% for LSTM. \\
    Traffic forecasting & LTC, LSTM & Mean Squared Error & LTC (0.099) is markedly lower than LSTM (0.169). \\
    Long-Range Arena (avg.) & Liquid-S4 & Accuracy (87.32\%) & Liquid-S4 reaches state-of-the-art performance. \\
    PhysioNet & CfC, ODE–RNN & Accuracy / AUC (similar) & CfC offers comparable accuracy but trains substantially faster. \\
    OCT image classification (retinal disease) & GLNN, LNN & Accuracy (\%), F1 score & GLNN achieves 0.98 Acc / 0.98 F1, surpassing LNN at 0.96 Acc / 0.88 F1. \\
    Noisy time-series prediction & UA-LNN, LNN, LSTM, MLP & $R^2$, RMSE, MAE & UA-LNN consistently outperforms all other models. \\
    Classic RNN benchmarks & LRCU, LSTM, GRU & Accuracy (\%) & LRCU outperforms both LSTM and GRU across tasks. \\
    Damped-sine trajectory prediction & GLNN, LNN, Neural ODE & Loss & GLNN (1.0738) significantly beats LNN (2.5494) and Neural ODE (1.9899). \\
    \bottomrule
  \end{tabularx}
\end{table*}

\subsection{Efficiency}
    \subsubsection{Memory efficiency: number of parameters, model size, and solving the “memory curse”}
        \begin{figure}[h]
            \centering
            \includegraphics[width=\linewidth]{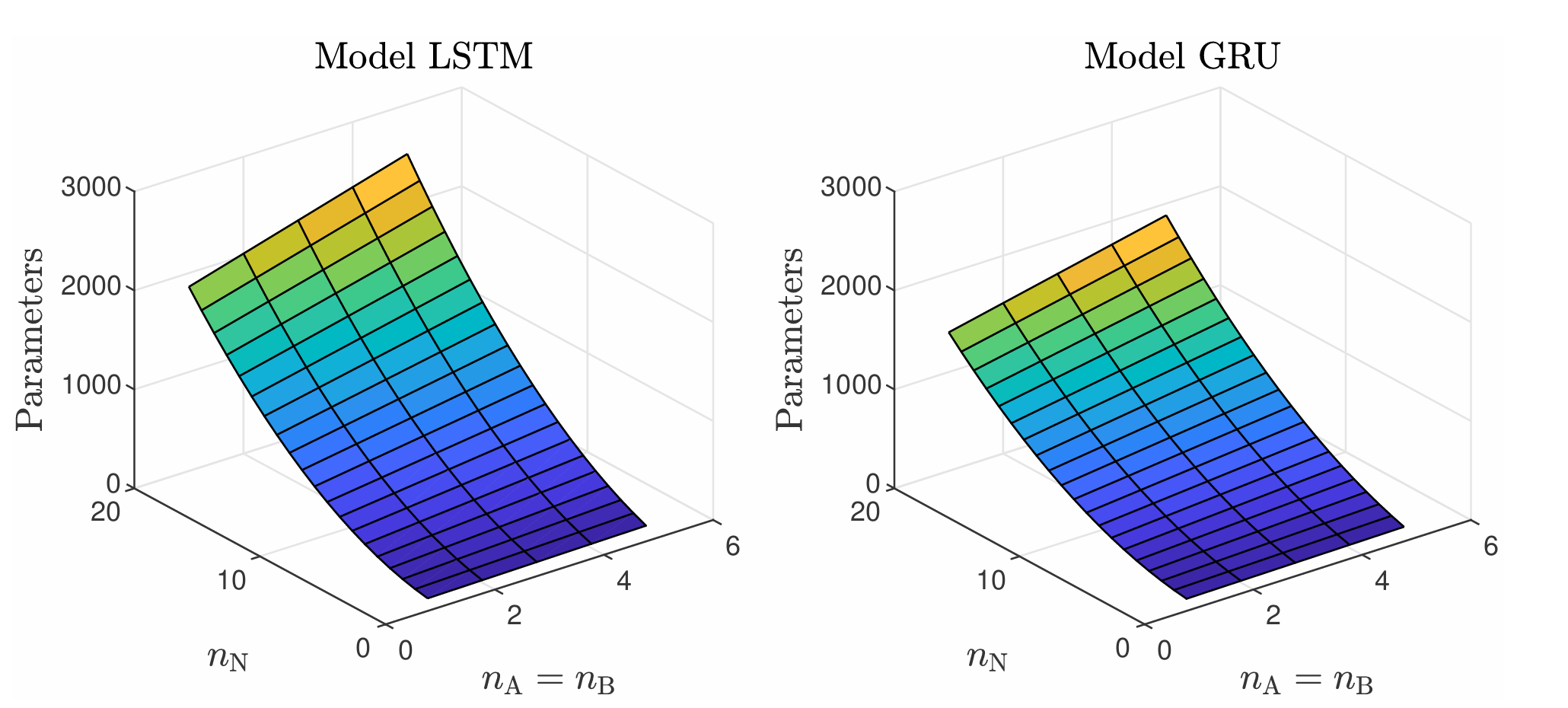}
            \caption{The number of the parameters of the Long Short-Term Memory (LSTM) and Gated Recurrent Unit (GRU) models as a function of the number of neurons and the order of the dynamics determined by \(n_A = n_P\).}
            \label{fig:s1}
        \end{figure}
        
\begin{figure}[h]
            \centering
            \includegraphics[width=\linewidth]{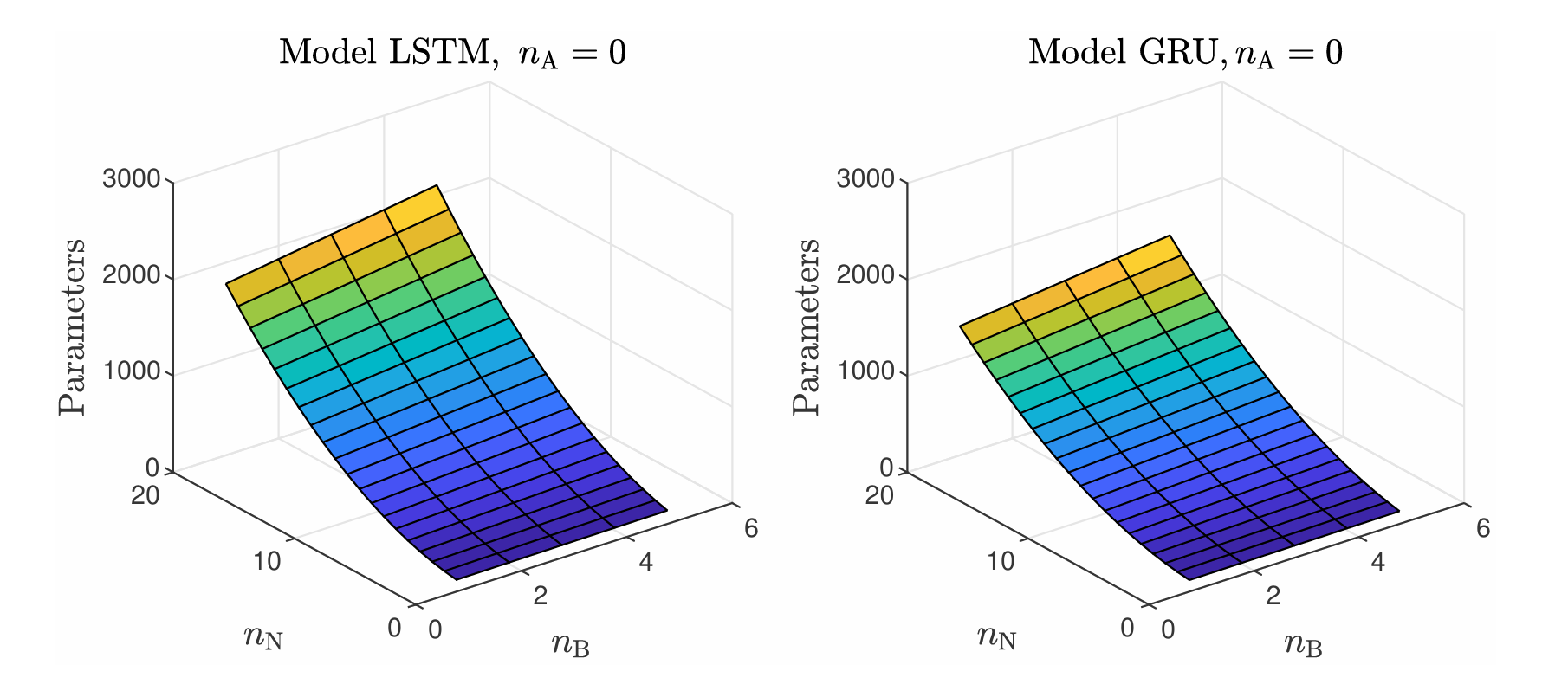}
            \caption{The number of parameters of the Long Short-Term Memory (LSTM) and Gated Recurrent Unit (GRU) models as a function of the number of neurons and the order of the dynamics determined by \(n_B\); \(n_A=0\).}
            \label{fig:s2}
        \end{figure}

        For RNN (LSTM/GRU), GRU typically has fewer parameters than LSTM due to its simpler gate structure, yet achieves comparable performance. For example, Figures~\ref{fig:s1} and ~\ref{fig:s2}, adapted from Lawryńczuk (2021) \cite{lawrynczuk2021lstm}, illustrate that GRU consistently has fewer parameters than LSTM under various configurations.

        For LNNs, NCPs are exceptionally compact; for example, an autonomous driving task uses only 19 neurons and 253 synapses, which is several orders of magnitude smaller than LSTM \cite{lechner2020neural}. CfC achieves state-of-the-art performance while using a “relatively small parameter set” \cite{hasani2022cfc}. A CfC model for IMDB(IMDB movie review dataset) has 75k parameters, while a CfC-NCP has 37k parameters.It has been reported that LNNs on Loihi-2 use 1–3 orders of magnitude fewer parameters than other NNs \cite{kaddoura2024exploring}. Liquid-S4 achieves state-of-the-art performance on speech commands with 30\% fewer parameters than S4 \cite{li2021curse}. The continuous-time nature and adaptive dynamics of LNNs may enable them to represent complex temporal patterns more effectively than discrete models that require many expansion steps or large hidden states to capture similar information.Efficiency-related indicators are summarized in Table~\ref{tab:overview-rnn-lnn}.
        
    \subsubsection{Computational efficiency: training speed, inference latency, and energy consumption (including neuromorphic implementations)}
        For RNNs (LSTM/GRU), GRU is typically faster than LSTM due to the lower computational complexity per unit. However, sequential processing limits the parallelization of both. Using these models for nonlinear optimization in MPC(Model Predictive Control (MPC)) involves significant computational complexity.
    
        For LNNs, a key advantage of CfC is speed. Their training/inference speeds are 1--5 orders of magnitude faster than their ODE-based counterparts, and they are much faster than LSTM due to the avoidance of ODE solvers \cite{hasani2022cfc}. For PhysioNet, CfC training speeds are 160 times faster than ODE-RNN \cite{hasani2022cfc}. Training LTC using vanilla BPTT (Backpropagation Through Time (BPTT)) may consume a significant amount of memory \((O(L×T))\) \cite{hasani2021ltc} . Their dependence on ODE solvers may make them slower than CfC. LRCU is an Eulerian discretization version of LRC, which is highly efficient. Training LRC with 1 expansion per step is 2.5 times faster than using 6 expansions \cite{farsang2024liquid}. The sparse and compact nature of NCP suggests high computational efficiency.On neuromorphic hardware (Loihi-2), LNN demonstrates exceptional efficiency \cite{kaddoura2024exploring}. LNN on Loihi-2 achieved 91.3\% accuracy for CIFAR-10 classification, consuming only 213 microjoules per frame.It has been reported that for such tasks, Loihi-2 achieves energy efficiency over 100 times higher than CPUs and nearly 30 times higher than GPUs. Compared to DNN/CNN/SNN on GPUs, LNN on Loihi-2 also exhibits lower latency (\(15.2 ms\)) and higher power efficiency (\(25.3 GOP/s/W\)).
    
        The drivers behind LNN's efficiency gains are multifaceted: some variants (CfC) optimize raw speed on traditional hardware by eliminating solvers, while others (NCP, universal LNN) exhibit exceptional energy efficiency and low parameter counts, making them suitable for edge devices and neuromorphic computing. This contrasts with RNNs, where efficiency improvements are typically incremental. RNNs (LSTM, GRU) inherently incur sequential processing costs and memory requirements, especially for long sequences (the memory curse) \cite{li2021curse}. LNNs address this issue from different angles: CfC improves speed by eliminating the ODE solver bottleneck \cite{hasani2022cfc}; NCP leverages extreme sparsity to achieve parameter and computational efficiency; and general-purpose LNNs show promise on energy-efficient neuromorphic hardware \cite{kaddoura2024exploring}. This suggests that LNNs are not merely aiming to be “better RNNs,” but are exploring fundamentally different efficiency pathways to adapt to different computational paradigms (fast traditional computing versus low-power neuromorphic computing).

\begin{table*}[!t]
  \caption{Efficiency-related Indicators Comparing Liquid Neural Networks (LNNs) and Recurrent Neural Networks (RNNs)}
  \label{tab:efficiency-indicators}
  \centering
  \footnotesize
  \begingroup
  \setlength{\tabcolsep}{4pt}
  \renewcommand{\arraystretch}{1.12}
  \setlength{\arrayrulewidth}{0.45pt}

  \begin{tabularx}{\textwidth}{|l|l|Y|Y|Y|Y|Y|}
    \hline
    \makecell{\textbf{Family}} & \makecell{\textbf{Model}} &
    \makecell{\textbf{\# Params}\\\textbf{(example / range)}} &
    \makecell{\textbf{Training}\\\textbf{Speed}} &
    \makecell{\textbf{Inference}\\\textbf{Latency}} &
    \textbf{Energy} &
    \makecell{\textbf{Key}\\\textbf{Efficiency Trait}} \\
    \hline

    \multicolumn{7}{|l|}{\textbf{Recurrent Neural Networks (RNNs)}} \\
    \hline
    RNN & GRU  & Typically $<$ LSTM & Faster than LSTM & Faster than LSTM & \NA & Simpler gating \\
    \hline
    RNN & LSTM & More than GRU      & \NA              & \NA              & \NA & Memory-heavy cells \\
    \hline

    \multicolumn{7}{|l|}{\textbf{Liquid Neural Networks (LNNs)}} \\
    \hline
    LNN & CfC  & IMDB 75k; NCP\mbox{-}CfC 37k & 16$\times$ faster than ODE\mbox{-}RNN & 1--5 orders faster & \NA & Solver-free closed form \\
    \hline
    LNN & NCP  & 19 neurons, 253 synapses     & \NA                              & \NA                & \NA & Ultra-compact sparse design \\
    \hline
    LNN & LTC  & \NA                          & \makecell{Solver-dependent;\\BPTT memory} & Solver-dependent & \NA & \NA \\
    \hline
    LNN & LRCU & \NA                          & Efficient (Euler discr.) & Efficient & \NA & Simplified LRC \\
    \hline
    LNN & Liquid-S4 & 30\% fewer than S4      & \NA                       & \NA       & \NA & Combines SSM efficiency \\
    \hline
    LNN & LNN (Loihi-2) & 1--3 orders fewer parameters & \NA & 15.2 ms (CIFAR-10) & 213 $\mu$J / frame & \makecell{Neuromorphic HW\\optimisation} \\
    \hline
  \end{tabularx}
  \endgroup
\end{table*}

\subsection{Generalization Ability}

    \subsubsection{Robustness to noise data and distribution changes}
    
        UA-LNN is specifically designed for noise resilience, modeling output uncertainty through Monte Carlo dropout \cite{akpinar2025novel}. They maintain excellent performance under strong noise in prediction and classification tasks (e.g., arrhythmia detection, cancer detection \cite{chahine2023robust}, \cite{akpinar2025novel}). NCP also demonstrates robustness; forward models that do not utilize temporal characteristics typically fail on noisy data, while NCP (derived from LTC) can filter out transient disturbances. LRC enhances generalization ability and accuracy, especially when using inexpensive solvers, and suppresses oscillations, contributing to stability \cite{farsang2024liquid}.
        
    \subsubsection{Out-of-distribution (OOD) generalization ability}
    
        One of the key advantages of LNNs lies in their OOD generalization capability. LNNs, especially in their differential equation and closed-form representations, demonstrate decision robustness when generalizing to new environments with drastic scene changes (e.g., flight navigation), which is a unique characteristic of these models \cite{chahine2023robust}. They learn to “extract task-relevant features and discard irrelevant ones.” Time series OOD is particularly challenging due to distribution shifts, diverse latent features, and non-stationary dynamics \cite{wu2025out}. The traditional independent and identically distributed (i.i.d.) assumption typically does not hold. A review on time series OOD organizes methodologies across three dimensions: data distribution, representation learning, and OOD evaluation. It emphasizes that LNNs leverage dynamic causal modeling to achieve adaptability and robustness \cite{wu2025out}.
        
    \subsubsection{The influence of continuous temporal dynamics and adaptability on generalization}
    
        The continuous-time nature of LNNs enables them to model system dynamics more faithfully, potentially capturing underlying causal structures that are invariant across distributions \cite{hasani2021ltc}, \cite{treven2023efficient}. Adaptive time constants and parameters enable LNNs to adapt to changing conditions during inference, which is crucial for OOD scenarios where test data differs from training data \cite{chahine2023robust}, \cite{kaddoura2024exploring}. This contrasts with static RNNs, where parameters are fixed after training. The superior OOD generalization ability of LNNs may stem from their ability to learn more fundamental and causal representations of tasks through continuous and adaptive dynamics.This makes them less susceptible to surface-level changes in the input distribution that may deceive models dependent on statistical correlations learned from a fixed training set. OOD generalization requires models to perform well on unseen data distributions. LNNs have been reported to perform well in this regard, for example, in flight navigation through “extracting tasks” and “discarding irrelevant features” \cite{chahine2023robust}. Their continuous-time dynamics and adaptability are key architectural features. These features may enable LNNs to learn the underlying causal mechanisms of systems rather than merely learning surface correlations present in the training data. Causal mechanisms are more likely to remain invariant across different distributions. Therefore, compared to models that overfit the characteristics of the training distribution, the architectural features of LNNs promote the learning of more robust and transferable representations, thereby achieving better OOD generalization.

\section{Case Study}

   This section conducts an empirical comparative study to evaluate the performance characteristics of liquid neural networks (LNNs) and standard recurrent neural networks (RNNs)—specifically long short-term memory networks (LSTMs) and gated recurrent units (GRUs)—on representative sequence data tasks. The case study draws on three different experimental settings: a trajectory prediction task using real-world motion capture data, a synthetic time series prediction task involving damped sine waves, and a high-dimensional complex prediction task involving ICU patient health states. 

    \subsection{Methodology}
    
        The primary objective of the case study is to compare LNN and traditional RNN side by side, focusing on their ability to learn time dependencies, efficiency in terms of model parameters and training duration, and accuracy in sequence prediction. To this end, we conducted the following experiments.
    
        The first experiment focuses on trajectory prediction using the Minari dataset. This dataset contains trajectories from simulated Walker2d agents, forming a complex, continuous-time, high-dimensional sequence modeling problem. The task is defined as predicting the next 17 observation features given the previous 10 states (each state includes 17 observation features and 6 action features). For this experiment, a Liquid Time Constant Network (LTC) with 64 hidden units was implemented. The dynamics of the LTC model are described by a system of ordinary differential equations, which are numerically solved using an adaptive solver. The input dimension of the LTC is 23 (observations + actions). As a baseline, a standard LSTM network was adopted, also configured with 64 hidden units and a single layer, and matched the input and output dimensions of the LTC model. Both models were trained for 20 cycles using the Adam optimizer with a learning rate of 0.001, and the mean squared error (MSE) between the predicted subsequent observations and the actual values was minimized as the loss function.
    
        The second experiment involves a synthetic time series prediction task, specifically modeling damped sine waves. This task was chosen to evaluate the model's ability to capture oscillatory and decaying patterns from a simpler, more controllable data source. To this end, a custom liquid neural network (LNN) inspired by neural circuit policy (NCP) principles and employing random neural wiring was developed. The LNN consists of 32 hidden neurons, whose ODE-based dynamics are integrated using the Euler method, with each input sample using 5 discrete time steps. The input and output of this task are both one-dimensional. A gated recurrent unit (GRU) network, also with 32 hidden units and a single layer, is implemented as the corresponding RNN model. In this experiment, both models were trained for 100 cycles using the Adam optimizer and MSE loss function, with a learning rate of 0.005 for the LNN and 0.01 for the GRU.

        The third experiment models Intensive Care Unit (ICU) patient state evolution on the MIMIC-III dataset using a CfC Liquid Neural Network and a GRU baseline, comparing accuracy, efficiency, and long-horizon robustness ~\cite{johnson2016mimic}. Data are discretized into non-overlapping 12-hour bins, aggregating physiological/lab features and interventions; the final design includes 18 physiological/laboratory variables and 3 intervention variables as predictors (interventions are not prediction targets). Preprocessing includes forward-fill of vitals, zero-fill for absent interventions, outlier capping/clamping to physiologically plausible ranges, and iterative imputation via Bayesian ridge. The CfC uses 2 layers with 128 hidden units; the GRU baseline has 2 layers with 128 hidden units. Both are trained with Adam (lr $10^{-3}$), batch size 64, for 30 epochs, minimizing MSE between $\hat{x}_{t+1}$ and $x_{t+1}$. Evaluation covers (i) single-step MAE/RMSE/$R^2$ in normalized feature space, (ii) $K$-step rollouts ($K\!=\!2,3,5$) with recursive predictions and true interventions each step, and (iii) efficiency: parameter counts, peak GPU memory during training, and throughput (examples/sec, steps/sec). Train/val/test is a 70/15/15 split by patient ID to avoid patient overlap. Lastly, both models were evaluated separately for robustness, where gaussian noise was introduced to the test data. This dataset and pipeline were adapted from related research by Lejarza et al., which used an alternate data-driven approach to model Patient Health: a stochastic Markov Decision Process (MDP) ~\cite{lejarza2021optimal}.

    \subsection{Experimental Results and Analysis}
        Many studies focus only on comparing the accuracy of LNNs and RNNs, while overlooking their computational efficiency~\cite{chahine2023robust}. The experiments conducted provide some comparative insights into the learning capabilities and efficiency of LNN and RNN models.and further details of efficiency-related indicators across different architectures are summarized in Table~\ref{tab:efficiency-indicators}.
        \vspace{1em}
        
        \subsubsection{Learning ability and prediction accuracy}
            \begin{figure}[t]
              \centering
              \begin{subfigure}[t]{0.32\linewidth}
                \includegraphics[width=\linewidth]{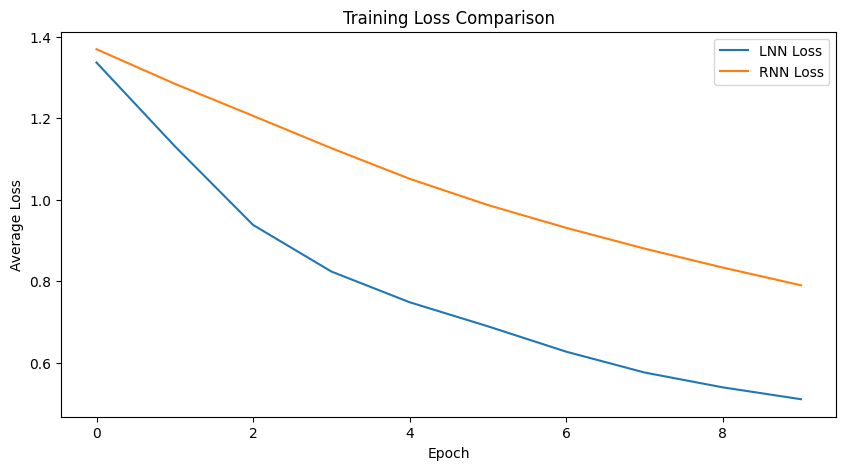}  
                \caption{Training loss comparison.}
                \label{fig:r3}
              \end{subfigure}
              \hfill
              \begin{subfigure}[t]{0.32\linewidth}
                \includegraphics[width=\linewidth]{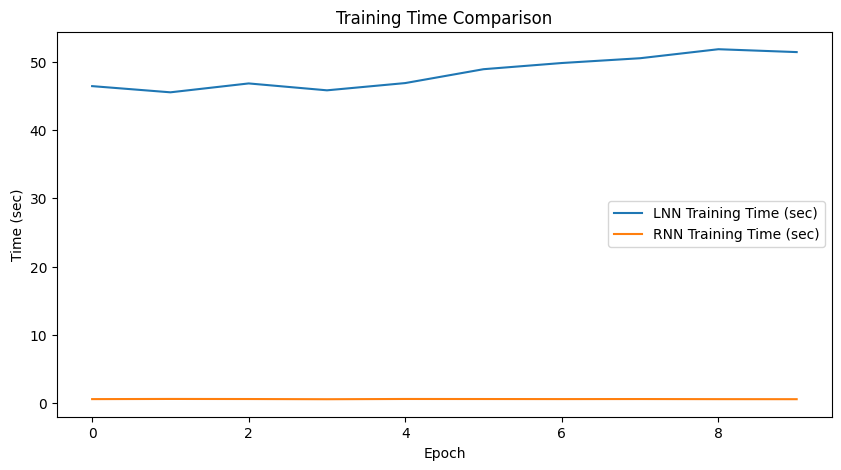}
                \caption{Comparison of training time per round.}
                \label{fig:r2}
              \end{subfigure}
              \hfill
              \begin{subfigure}[t]{0.32\linewidth}
                \includegraphics[width=\linewidth]{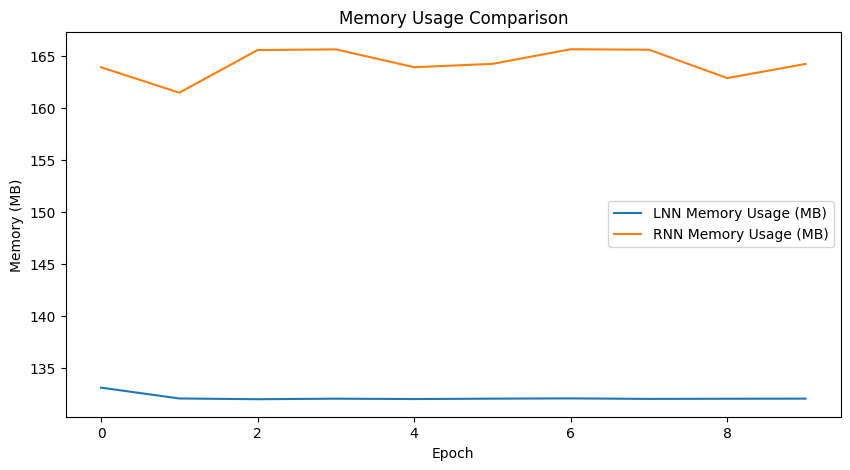}
                \caption{Comparison of training memory usage.}
                \label{fig:r1}
              \end{subfigure}
              \caption{Comparison of (a) average loss, (b) training time per round, and (c) training memory usage between Liquid Time-Constant (LTC) (Liquid Neural Network (LNN)) and Long Short-Term Memory (LSTM) (Recurrent Neural Network (RNN)) in the Walker2d trajectory prediction task.}
              \label{fig:r1-3}
            \end{figure}
            
\begin{figure}[h]
                \centering
                \includegraphics[width=\linewidth]{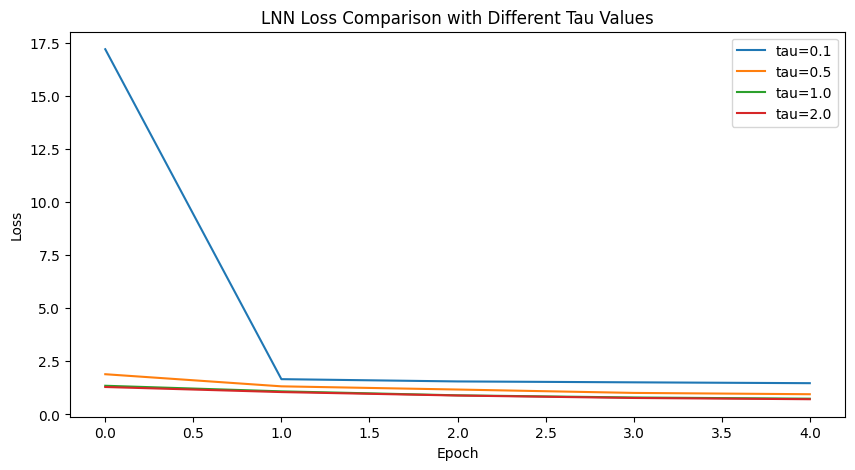}
                \caption{True values and Liquid Time-Constant (LTC) (Liquid Neural Network (LNN)) and Long Short-Term Memory (LSTM) (Recurrent Neural Network (RNN)) prediction results for a specific feature of the Walker2d trajectory.}
                \label{fig:walker-feature}
            \end{figure}

            In terms of learning ability and prediction accuracy, both types of networks demonstrated the ability to learn complex patterns from sequence data. For the Walker2d trajectory prediction task, comparing the average loss curves of LTC (labeled as LNN in the experiment) and LSTM (labeled as RNN in the experiment) (as shown in Figure~\ref{fig:r3}), it can be seen that in the early stages of training, the losses of both models decreased rapidly, indicating that they were able to effectively learn the dynamic characteristics of the data. Although LSTM appears to reach a lower loss plateau faster and maintain slightly lower loss values throughout the entire process in this specific run, LTC also exhibits a trend of continuous learning and loss reduction. In the prediction of individual trajectory features (as shown in Figure ~\ref{fig:walker-feature}), both LTC and LSTM achieve varying degrees of accuracy in approximating the true trajectory, which intuitively reflects their predictive capabilities.
            
\begin{figure}[h]
                \centering
                \includegraphics[width=\linewidth]{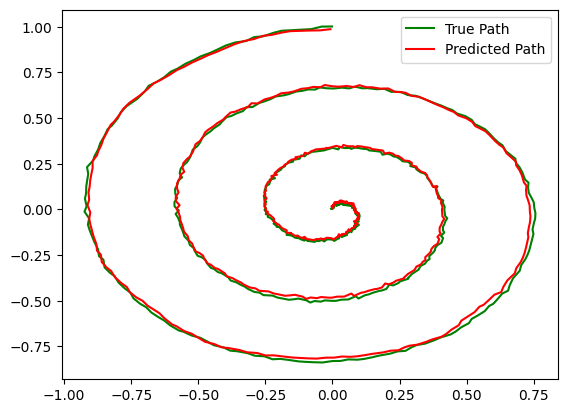}
                \caption{LNN fitting of a damped sine wave (ground truth vs.\ LNN prediction). The LNN closely tracks both the periodic component and the decaying envelope.}
                \label{fig:r7}
            \end{figure}

\begin{figure}[h]
                \centering
                \includegraphics[width=\linewidth]{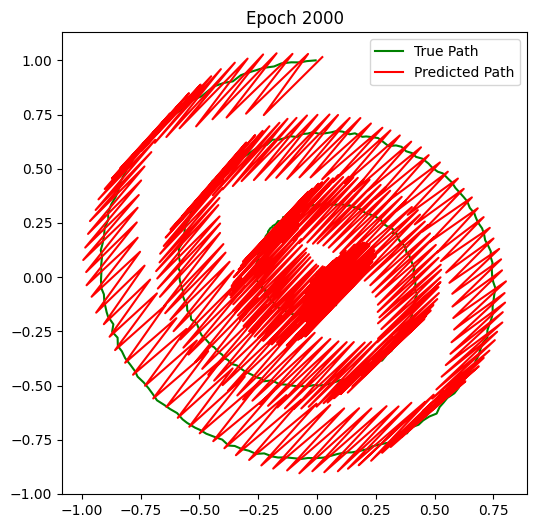}
                \caption{GRU fitting of a damped sine wave (ground truth vs.\ GRU prediction).
  The GRU learns the overall trend but shows larger deviations near turning points 
  and in following the decaying envelope.}
                \label{fig:r8}
            \end{figure}

            For the synthetic damped-sine-wave prediction task, the learned trajectories in 
            Figs.~\ref{fig:r7} and \ref{fig:r8} compare a custom LNN with a GRU. 
            Both models capture the periodicity and the decaying envelope, reducing the prediction error; 
            however, the LNN yields a visibly tighter fit across the horizon—especially near peaks and 
            zero-crossings—indicating a more faithful modeling of the underlying damped dynamics. 
            Under additive zero-mean Gaussian input noise ($\sigma/\text{amp}=0.10$), both models attempt 
            to recover the clean waveform, with the LNN exhibiting stronger smoothing and noise rejection. 
            A more comprehensive robustness comparison would require quantitative metrics and multiple 
            noise settings, but these observations suggest that the LNN architecture is better suited to 
            this noisy synthetic task.

\begin{figure}[h]
                \centering
                \includegraphics[width=\linewidth]{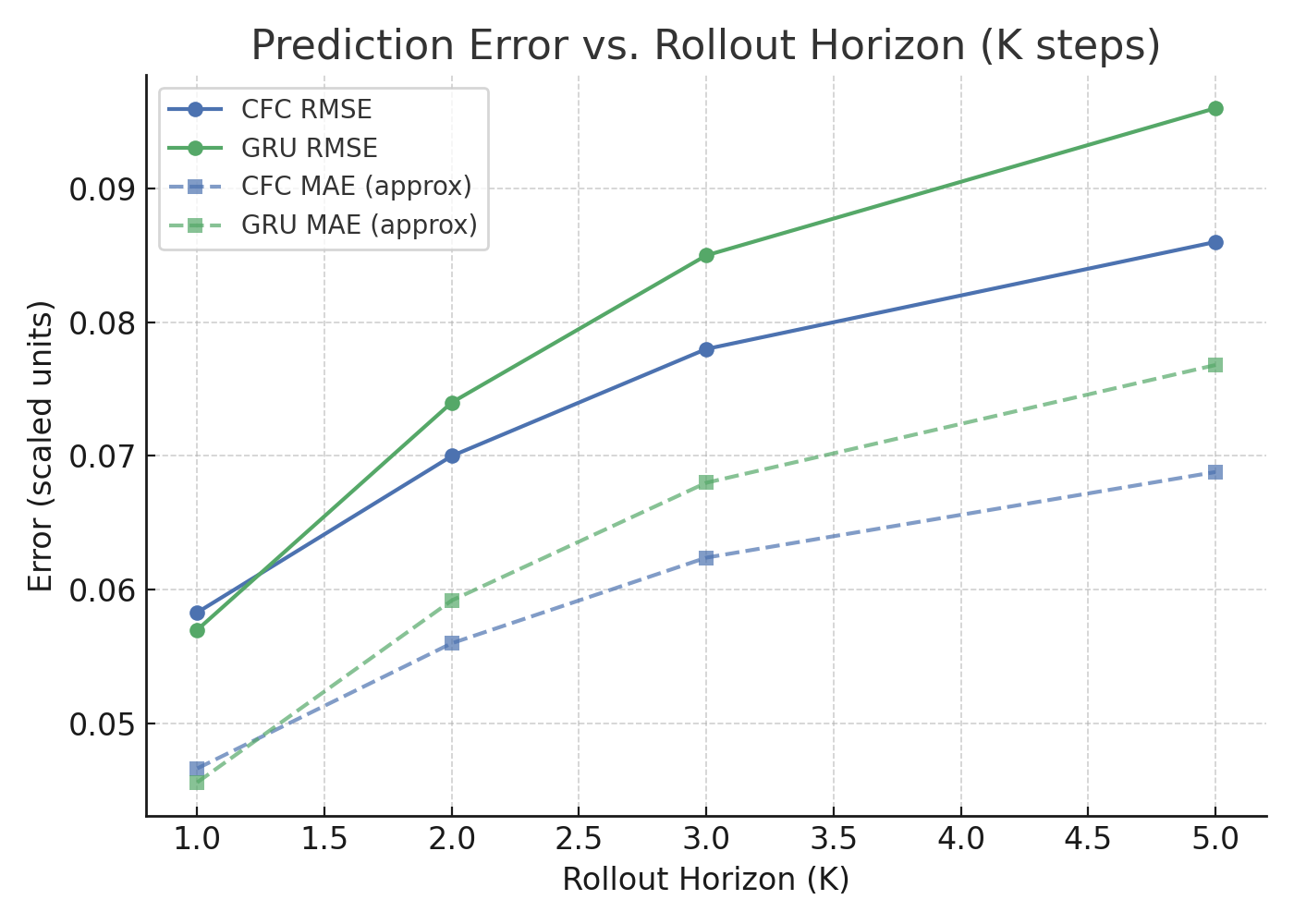}
                \caption{Prediction error comparison (RMSE and MAE) between Closed-form Continuous-time (CfC) and GRU across single-step and multi-step rollouts ($K=2,3,5$) on ICU patient trajectories.}
                \label{fig:icu-rollout-error}
            \end{figure}

            The ICU Patient experiment also found promising results in relation to CfC. As seen in Figure ~\ref{fig:icu-rollout-error}, the single-step precision for CfC and GRU is effectively identical (MAE/RMSE differences of the order of $10^{-3}$ in normalized space), indicating that there is no short-horizon loss due to the compactness of CfC. Under multi-step rollouts, CfC exhibits consistently lower error accumulation, achieving \(\approx\)10--11\% lower RMSE by \(K{=}5\), suggesting more stable long-horizon dynamics. 
            \vspace{1em} 
            
            \begin{figure}[h]
                \centering
                \includegraphics[width=\linewidth]{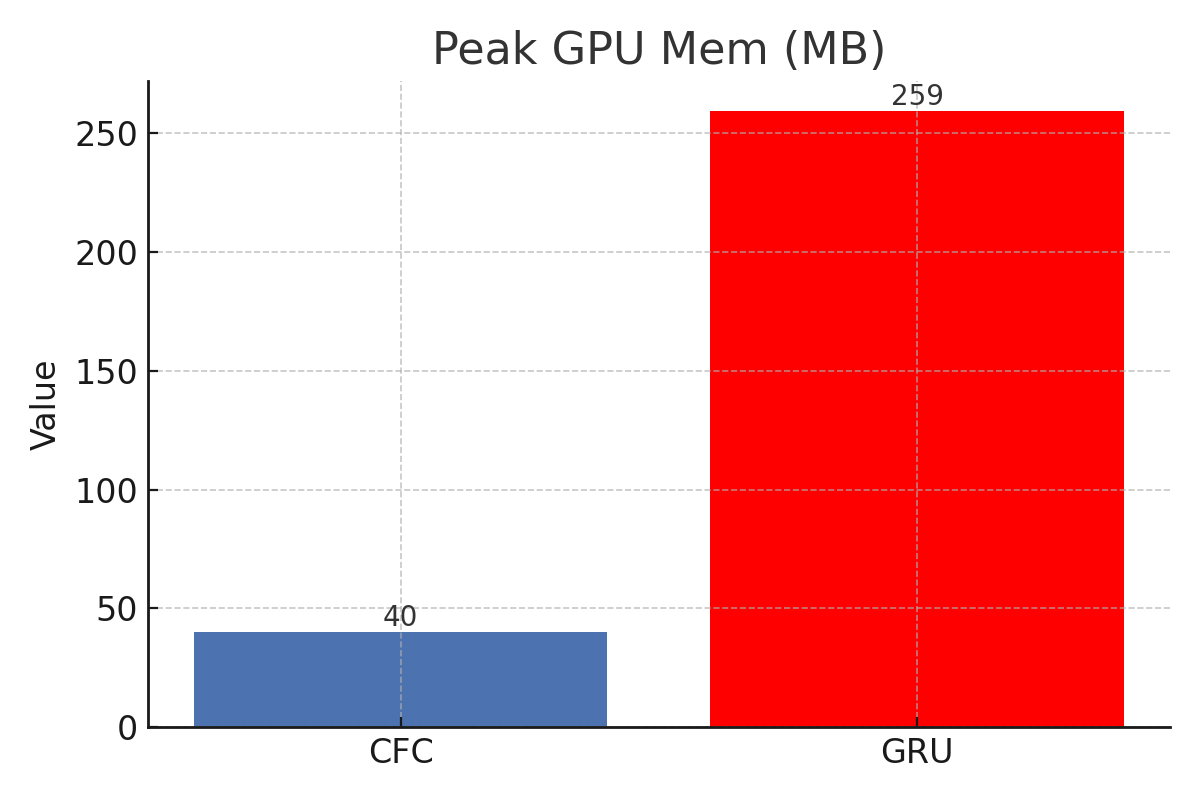}
                \caption{Peak GPU memory usage during training for CfC and GRU models on ICU patient trajectory prediction.}
                \label{fig:icu-mem-usage}
            \end{figure}

        \subsubsection{Efficiency}
            
        We evaluate robustness under additive i.i.d. Gaussian noise; unless otherwise noted, noise is zero-mean with variance chosen to yield a moderate signal-to-noise ratio.
        \vspace{-1em} 
        \begin{figure}[t]
              \centering
              \begin{subfigure}[t]{0.48\linewidth}
                \includegraphics[width=\linewidth]{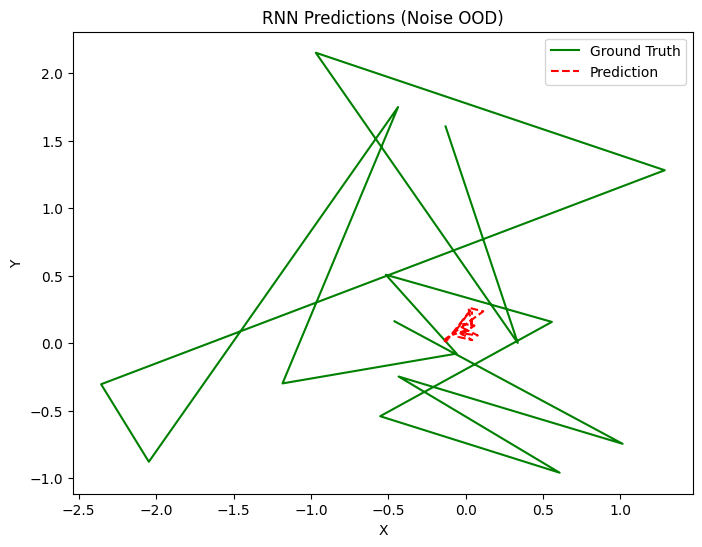}
                \caption{Long Short-Term Memory (LSTM) (Recurrent Neural Network (RNN)) prediction on noisy sequences (additive zero-mean Gaussian noise).}
                \label{fig:r5}
              \end{subfigure}
              \hfill
              \begin{subfigure}[t]{0.48\linewidth}
                \includegraphics[width=\linewidth]{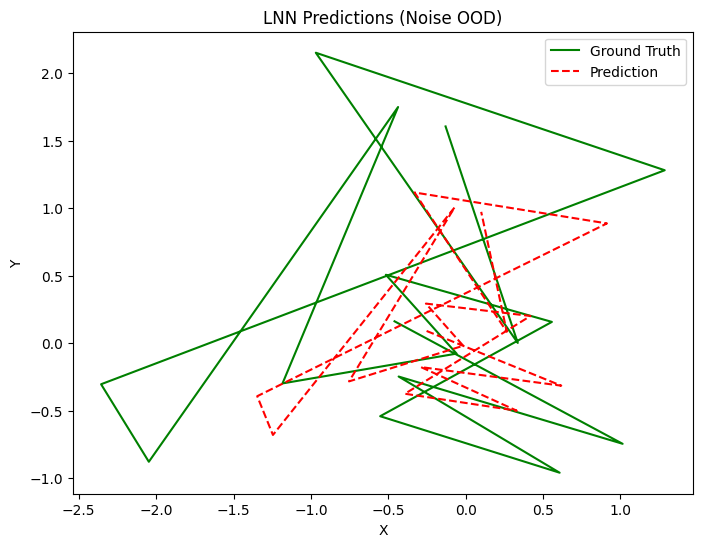}
                \caption{Liquid Time-Constant (LTC) (Liquid Neural Network (LNN)) prediction on noisy sequences (additive zero-mean Gaussian noise).}
                \label{fig:r6}
              \end{subfigure}
              \caption{(a) Long Short-Term Memory (LSTM) (Recurrent Neural Network (RNN)) and (b) Liquid Time-Constant (LTC) (Liquid Neural Network (LNN)) model prediction performance on noisy sequences (additive zero-mean Gaussian noise) (including true signal, noise input, and model prediction).}
              \label{fig:r5-6}
            \end{figure} 
            
          In terms of efficiency, the experiments revealed significant differences between different architectures. The comparison of training times in the Walker2d trajectory prediction task (as shown in Figure~\ref{fig:r2}) clearly indicates that the LTC model using an ODE solver requires longer computation time per training epoch (e.g., approximately 7–8 seconds per epoch for LTC and approximately 1–2 seconds per epoch for LSTM). This reflects the higher computational overhead typically associated with continuous-time dynamic modeling. However, in terms of memory usage (as shown in Figure~\ref{fig:r1}), LTC exhibits relatively stable and slightly lower GPU memory consumption compared to LSTM during training. Combining the parameter count information obtained from torchinfo.summary in the previous analysis (in this experimental setup, the LTC model with 64 hidden units has approximately 10k-12k parameters, while the LSTM model has approximately 20k–22k parameters), LTC demonstrates an advantage in parameter efficiency. This suggests that LNNs may achieve effective encoding of complex dynamics with fewer parameters, though this may come at the cost of longer training time per epoch. For custom LNNs and GRUs on the damped sine wave task, although direct comparisons of parameter counts and training times are not shown, GRUs are generally considered more efficient than LSTM, while the efficiency of custom LNNs highly depends on specific implementation details such as the number of neurons, connection density, and ODE integration step size. For the high-dimensional ICU patient health prediction task, CfC uses \(\sim\)18\(\times\) fewer parameters (0.011M vs.\ 0.203M) and \(\sim\)6.5\(\times\) lower peak GPU memory (40 MB vs.\ 259 MB) during training, but trains \(\sim\)3\(\times\) slower in throughput terms (e.g., 384 vs.\ 1298 examples/sec) (Figure ~\ref{fig:icu-mem-usage}). Overall, the ICU findings complement the earlier case studies: CfC preserves short-horizon accuracy while improving long-horizon robustness and resource efficiency, with a practical trade-off in training speed.
          \vspace{1em} 
            
        \subsubsection{General observation and implicit generalization abilities}

            In terms of general observation and implicit generalization capabilities, the LTC model demonstrates strong learning performance on complex, high-dimensional Walker2d trajectory data, combined with its parameter efficiency, suggesting its potential for tasks requiring precise modeling of continuous dynamic systems. Its continuous-time nature theoretically facilitates handling sequences with irregular sampling or time-varying dynamics. Performance under noisy conditions (as shown in Figure ~\ref{fig:r5} and Figure ~\ref{fig:r6}) also preliminarily demonstrates LNN's ability to resist interference and extract signal essence to some extent, while the RNN baseline exhibits more severe performance degradation under the same noisy conditions, which is crucial for enhancing model robustness and generalization to noisy real-world data. 
            
            Although this case study did not directly conduct rigorous out-of-distribution (OOD) testing, the inherent characteristics of the LNN architecture, such as continuous-time processing, adaptive dynamics, and parameter efficiency demonstrated in certain variants, are all favorable factors for improving generalization performance. The model's relatively small number of parameters typically implies lower overfitting risk, which may indirectly promote generalization.

            For the ICU patient case study, gaussian noise was injected into the evaluation data after training to assess stability under measurement uncertainty. For each rollout horizon $K \in \{1,2,3,5\}$, noise scales $\sigma \in \{0.0,0.01,0.02,0.05\}$ were tested, with $\sigma$ defined relative to the min–max normalized feature range. Models were rolled forward for $K$ steps with noisy inputs, while true interventions were retained, and mean absolute error (MAE) and root mean squared error (RMSE) were computed against the ground truth. 

\begin{figure}
                \centering
                \includegraphics[width=\linewidth]{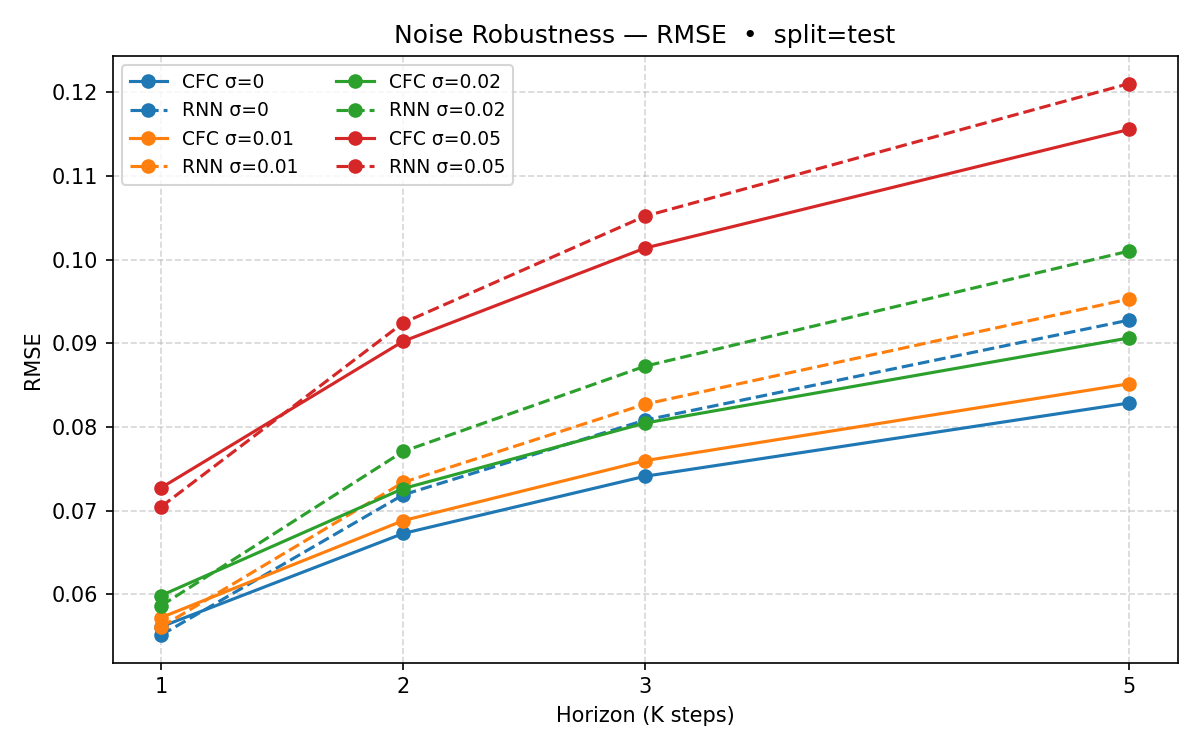}
                \caption{Robustness comparison of CfC and RNN-GRU (RMSE).}
                \label{fig:ICU_robustness}
            \end{figure}

            As seen in figure~\ref{fig:ICU_robustness}, Both models tolerate small perturbations ($\sigma=0.01$–$0.02$) with only $\sim$5–15\% degradation in RMSE. At $\sigma=0.02$, CfC and GRU behave similarly, showing $\sim$13–15\% higher MAE and $\sim$6–10\% higher RMSE compared to the baseline. At stronger noise ($\sigma=0.05$), differences emerge: CfC exhibits a $\sim$50\% increase in MAE and $\sim$40\% increase in RMSE, while GRU increases by $\sim$47\% MAE and $\sim$31\% RMSE. Across rollout horizons, relative degradation is consistent (e.g., $K=1$ vs.\ $K=5$), indicating that noise effects do not amplify substantially with longer rollouts. In this experiment, CfC and GRU are comparably robust under mild perturbations. 

        In summary, this case study provides empirical evidence for the comparison between LNN and RNN through concrete experimental results. The results indicate that LNN can effectively learn complex sequence dynamics and demonstrate potential in terms of parameter efficiency, although certain implementations (such as LTC, which relies on complex ODE solvers) may be slower in training compared to traditional RNNs. These observations provide valuable insights into the intrinsic characteristics of different neural network architectures when handling sequence data and the trade-offs in practical applications.
        
\section{Future Work and Open Challenges}
    \subsection{Enhancing the scalability of liquid neural networks}
        Although certain liquid models such as Neural Circuit Policies (NCPs) are notably compact \cite{lechner2020neural}, scaling liquid neural networks (LNNs) to very large datasets and high-dimensional state spaces remains open. Promising directions include (i) algorithmic improvements to stiff/fast solvers and stability-aware integration, (ii) solver-free formulations that preserve continuous-time benefits while reducing overheads (e.g., Closed-form Continuous-time models, CfC \cite{hasani2022cfc}), and (iii) memory-efficient training alternatives to BPTT together with distributed and mixed-precision training. On the systems side, mapping continuous dynamics to parallel hardware requires careful partitioning and scheduling, and exploiting sparsity and event-driven computation at scale.
        
    \subsection{Advancing Robustness and OOD Generalization in Dynamic Environments}
        A central challenge is adaptation under distribution shift and temporal non-stationarity. Future work should pursue online/continual learning procedures for LNNs and RNNs, stronger uncertainty quantification and calibration (extending the UA-LNN line \cite{akpinar2025novel}), and invariance-inducing objectives that maintain performance under sensor noise, missing data, or regime changes. Robust control viewpoints and OOD navigation results for liquid models suggest promising headroom, but stress testing in rapidly changing environments is still required \cite{chahine2023robust}.
        
    \subsection{Optimizing LNNs for Specialized Hardware and Edge Computing}
        Liquid models are a natural fit for low-power, event-driven, and neuromorphic substrates. Co-design of algorithms and hardware—quantization, pruning, low-rank/state-space compression, and operator fusion—can reduce latency and energy while preserving stability guarantees. Solver-free liquid variants are particularly attractive for embedded deployment \cite{hasani2022cfc}. A systematic evaluation protocol (accuracy–latency–energy–memory) across edge devices will help identify when liquid dynamics provide the strongest advantage.
        
    \subsection{New Applications and Hybrid Methods}
        It is valuable to explore how LNNs can be applied beyond current use cases, including modeling complex physical processes, advanced control, and clinical decision support. Hybrid architectures that couple continuous-time liquid dynamics with complementary inductive biases—such as Transformers for long-range attention or graph neural networks for relational structure—may combine the strengths of each paradigm.
    \subsection{Synthesis}
        Closing the gap between the attractive theoretical properties of liquid dynamics (continuity, adaptivity, stability) and reliable large-scale deployment will likely hinge on three threads: scalable training and solver design (including solver-free approaches \cite{hasani2022cfc}), principled robustness with calibrated uncertainty under shift (building on UA-LNN \cite{akpinar2025novel} and robustness studies \cite{chahine2023robust}), and hardware–algorithm co-design for efficient inference at the edge. Consolidated theory (expressivity, identifiability, and stability under discretization) together with open, noise-aware benchmarks will provide the foundations for the next wave of liquid models \cite{hasani2021ltc}, \cite{lechner2020neural}.
        
    \subsection{Integrating CfC Models into Policy Optimization Frameworks}

        Another important extension concerns bridging CfC-based patient trajectory models with policy optimization for clinical decision-making. Lejarza et al. \cite{lejarza2021optimal} previously formulated ICU discharge planning as a finite Markov decision process (MDP), where policy iteration over handcrafted, discretized patient states yielded stable discharge policies. While effective, this approach was constrained by the need for a manually specified state space and limited flexibility in representing continuous physiological variables.
        
        Our case study suggests that replacing the fixed MDP transition dynamics with a learned CfC model could create a more adaptive simulation environment for reinforcement learning. In such a framework, the CfC serves as the generative dynamics model, providing realistic patient state transitions under clinical interventions. Policy iteration or other dynamic programming methods could then operate on this learned environment to identify discharge policies that balance patient safety and resource efficiency. This integration would merge the robustness and parameter efficiency of liquid neural networks with established policy optimization techniques, potentially yielding more flexible and clinically relevant decision-support tools. Developing and validating this combined approach represents a promising direction for future research.

\section{Conclusion}
    A comparative analysis of LNNs and RNNs reveals the advantages of LNNs in handling continuous-time dynamics, adaptability, and OOD generalization and specialized hardware efficiency, as confirmed by numerous studies. In contrast, while RNNs possess mature capabilities, they also have limitations such as the “memory curse” and challenges in handling truly continuous processes.
    
    LNN offers an intriguing avenue for overcoming several fundamental limitations of traditional RNNs. Its bio-inspired, ODE-based framework provides a richer foundation for modeling complex dynamic systems.
    
    The field is evolving rapidly, with a clear trend toward more adaptive, efficient, and robust neural architectures. LNN is at the forefront of this movement, particularly for applications that require continuous adaptation and interaction with the physical world. The synergies with neuromorphic computing are particularly noteworthy and may drive future innovations. The development of more specialized LNN variants indicates that the field is maturing, offering a diverse toolkit for sequence modeling. The rise of LNNs is not merely an incremental improvement over RNNs; it represents a potential paradigm shift toward neural architectures that are more inherently aligned with the continuous and dynamic characteristics of many real-world problems, offering a principled approach to building smarter and more adaptive systems.

\bibliographystyle{IEEEtran}
\bibliography{references}

\end{document}